\documentclass[10pt,twocolumn,letterpaper]{article}

\usepackage{cvpr}              

\definecolor{cvprblue}{rgb}{0.21,0.49,0.74}
\usepackage[pagebackref,breaklinks,colorlinks,allcolors=cvprblue]{hyperref}

\def\paperID{****} 
\def\confName{CVPR}
\def\confYear{2025}
\usepackage[utf8]{inputenc} 
\usepackage[T1]{fontenc}    
\usepackage{hyperref}       
\usepackage{url}            
\usepackage{booktabs}       
\usepackage{amsfonts}       
\usepackage{nicefrac}       
\usepackage{microtype}      
\usepackage{xcolor}         
\usepackage{times}

\usepackage{microtype}
\usepackage{lipsum}
\usepackage{graphicx}
\usepackage{subcaption}
\usepackage{sidecap}
\usepackage{caption}
\usepackage{booktabs} %
\usepackage{amsfonts}       %
\usepackage{nicefrac}       %
\usepackage{microtype}      %
\usepackage{comment}
\usepackage{enumitem}
\usepackage{wrapfig,tikz}
\usepackage{amsmath,longtable,fancyhdr}
\usepackage{bigstrut, tabularx, multirow, makecell, diagbox}
\usepackage[export]{adjustbox}
\usepackage{graphicx,float}
\usepackage{amssymb,amsthm, mathtools}
\usepackage{stackrel}
\usepackage{color}
\usepackage{colortbl}
\usepackage{theoremref}
\usepackage{cases}
\usepackage{stmaryrd}
\usepackage{mathabx}
\usepackage{dsfont}
\usepackage{xcolor}
\usepackage{algorithmic}
\usepackage{algorithm}
\usepackage{bm}
\usepackage{footmisc}

\newcolumntype{C}[1]{>{\centering\arraybackslash}m{#1}}
\newcolumntype{P}[1]{>{\centering\arraybackslash}p{#1}}

\usepackage{mdframed}

\makeatletter
\usepackage{xspace} 
\def\@onedot{\ifx\@let@token.\else.\null\fi\xspace}
\DeclareRobustCommand\onedot{\futurelet\@let@token\@onedot}

\usepackage{amsmath,amsfonts,bm}

\def\eqref#1{equation~\ref{#1}}

\def\1{\bm{1}}

\DeclareMathAlphabet{\mathsfit}{\encodingdefault}{\sfdefault}{m}{sl}
\SetMathAlphabet{\mathsfit}{bold}{\encodingdefault}{\sfdefault}{bx}{n}

\title{StyleRWKV: High-Quality and High-Efficiency Style Transfer \\ with RWKV-like Architecture}

\author{
    Miaomiao Dai\footnotemark[1], 
    Qianyu Zhou\thanks{\textit{These authors contributed equally.}}, 
    Lizhuang Ma\textsuperscript{†} \\
    Shanghai Jiao Tong University \\
    {\tt\small \{dmm2020, zhouqianyu, lzma@\}sjtu.edu.cn}
}

\begin{document}
\maketitle
\begin{abstract}
    Style transfer aims to generate a new image preserving the content but with the artistic representation of the style source. Most of the existing methods are based on Transformers or diffusion models, however, they suffer from quadratic computational complexity and high inference time. RWKV, as an emerging deep sequence models, has shown immense potential for long-context sequence modeling in NLP tasks. In this work, we present a novel framework StyleRWKV, to achieve high-quality style transfer with limited memory usage and linear time complexity. Specifically, we propose a Recurrent WKV (Re-WKV) attention mechanism, which incorporates bidirectional attention to establish a global receptive field. Additionally, we develop a Deformable Shifting (Deform-Shifting) layer that introduces learnable offsets to the sampling grid of the convolution kernel, allowing tokens to shift flexibly and adaptively from the region of interest, thereby enhancing the model's ability to capture local dependencies. Finally, we propose a Skip Scanning (S-Scanning) method that effectively establishes global contextual dependencies. Extensive experiments with analysis including qualitative and quantitative evaluations demonstrate that our approach outperforms state-of-the-art methods in terms of stylization quality, model complexity, and inference efficiency.
\end{abstract}
\vspace{-0.5cm}    
\section{Introduction}
\begin{figure}[htbp]
	\begin{center}
		\centerline{\includegraphics[width=0.5\textwidth]{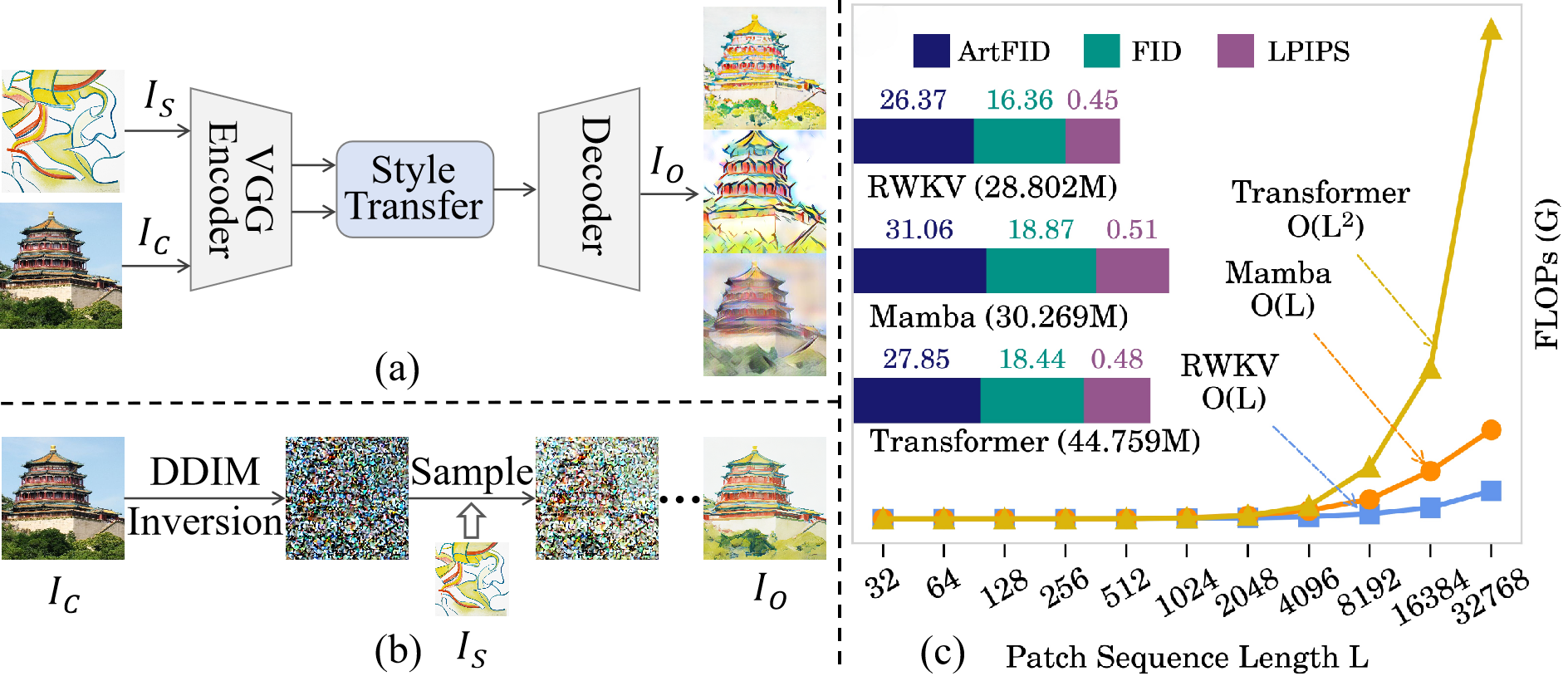}}
		\caption{
       Comparison of existing neural style transfer methods. (a) Traditional (CNN-based or attention-based) methods suffer from limited receptive field or quadratic
computational complexity, leading to unsatisfactory results. (b) Diffusion-based methods rely on multiple iterative denoising steps, making them significantly time-intensive. (c)  Performance and efficiency comparison. (Left) Overall stylization performance acquired by different methods with relative parameters, (Right) FLOPs increase with sequence length.}
		\label{figure1}
	\end{center}
        \vspace{-1.0cm}
\end{figure}
Neural Style Transfer (NST)~\cite{b1, b5, b6, b8, b9, b10, b11, b12, b13, b14, b15, b16, b50, b51, b52, b53, b59, b60, b61, b62, b63, b64, b65, b66, b69, b71} aims to generate a new image which has the content (\emph{e.g.,} objects, layout) of a given content image and the style (\emph{e.g.,} color or texture) of another style image. Most of the existing methods are based on Transformers~\cite{b18, b70} or Diffusion Models (DMs)~\cite{b19, b21, b53}, which often suffer from high computation burden. Particularly, Transformer-based models~\cite{b11, b54, b55} are constrained by quadratic complexity, which hampers their scalability for high-resolution image processing. Meanwhile, DMs-based methods \cite{b13, b14, b15, b16, b58} are hindered by their time-consuming generation process, requiring multiple iterative denoising steps for inference.

Recent advancements in Mamba \cite{b22, b24, b25, b67, b68, b69} demonstrate promising capabilities in modeling long-range dependency with linear complexity. Mamba-based models~\cite{b24, b25} are recently introduced in style transfer to achieve both linear complexity and low memory consumption. However, they cannot cope well with high resolution images \emph{i.e.,} long patch sequence, due to the auxiliary computation operations and specific optimization strategy in its linear attention mechanism, \emph{e.g.,} global modeling, feature fusion, context enhancement, \emph{etc.}, which will exhibit a higher computational load, especially in long sequences. Thus, a natural question is that \textit{can we design a neural style transfer method with linear complexity that achieves a trade-off between stylization quality (in both the low- and high-resolution images) and efficiency
(i.e., faster inference speeds and lower memory usage)?}
Such a challenging problem still remains unexplored in NST.

We notice the recent advance of RWKV~ \cite{b23, b27, b28, b29, b31, b32, b34} model within natural language processing (NLP), serving as a promising solution for long-context sequence modeling. As shown in Fig.~\ref{figure1}~(c), we visualize the stylization performance and efficicency of various NST methods. We observe that Transformer-based methods suffer from quadratic complexity as the patch sequence become longer. Although Mamba achieve linear complexity, it is inferior to RWKV especially in long sequences. In contrast, RWKV-based style transfer exhibit higher inference efficiency than Mamba-based method for long sequences (L\textgreater2048), and lower model complexity (28.802M) and superior stylization quality compared to both Transformer- and Mamba-based counterparts.

Motivated by the above analysis, we introduce StyleRWKV, a novel style transfer framework that excels in linear computation complexity, limited memory usage and high efficiency in not only low resolution but also high resolution images. Specifically, StyleRWKV uses an RWKV-like architecture to model the global features of both content and style images through a stack of our presented Style Transfer RWKV (ST-RWKV) blocks. To capture the semantic dependencies between local and global features, the architecture employs a multi-stage hierarchical design that progressively encodes multi-scale image features. Within each ST-RWKV block, we introduce three critical core components: Recurrent WKV (Re-WKV) attention mechanism, Deformable Shifting (Deform-Shifting) layer, Skip Scanning (S-Scanning) mechanism.
Firstly, we introduce Re-WKV Attention, which efficiently capture global dependencies with linear complexity. Unlike the unidirectional causal WKV attention in the original RWKV~\cite{b23}, which restricts the receptive field, Re-WKV recurrently employs a bidirectional attention to achieve a global receptive field, enabling more comprehensive context modeling. Secondly, we propose Deform-Shifting layer, which enhances the modeling of local dependencies by dynamically shifting tokens from regions of interest (ROIs). Unlike previous token shifting layers that use simple interpolation to shift adjacent tokens in fixed directions~\cite{b23}, Deform-Shifting improves the interaction between neighboring tokens within an image's object structure by adaptively shifting them based on the region of interest. Finally, a Skip Scanning (S-Scanning) mechanism is integrated into the Re-WKV, effectively capturing both local and global dependencies. Using the atrous sampling strategy ensures critical context is preserved without compromising computational efficiency.

Our contributions can be summarized into three aspects:
\begin{itemize}
\item We propose StyleRWKV, a novel RWKV-like model for style transfer tasks, offering efficient linear time complexity, low memory usage and global receptive fields, for both low-resolution and high-resolution images.

\item We introduce a Recurrent WKV attention that efficiently captures global dependencies with linear complexity. Besides, we propose a Deformable Shifting module, which dynamically shifts tokens from regions of interest to more accurately capture local dependencies within an image's object structures. Additionally, Skip Scanning mechanism is designed to enhance its ability to model distant contextual dependencies effectively.

\item Extensive experiments demonstrate its effectiveness and superiority over state-of-the-art methods in terms of stylization quality, inference time, and memory usage.
\end{itemize}
\begin{figure*}[htbp]
	\begin{center}
		\centerline{\includegraphics[width=170mm]{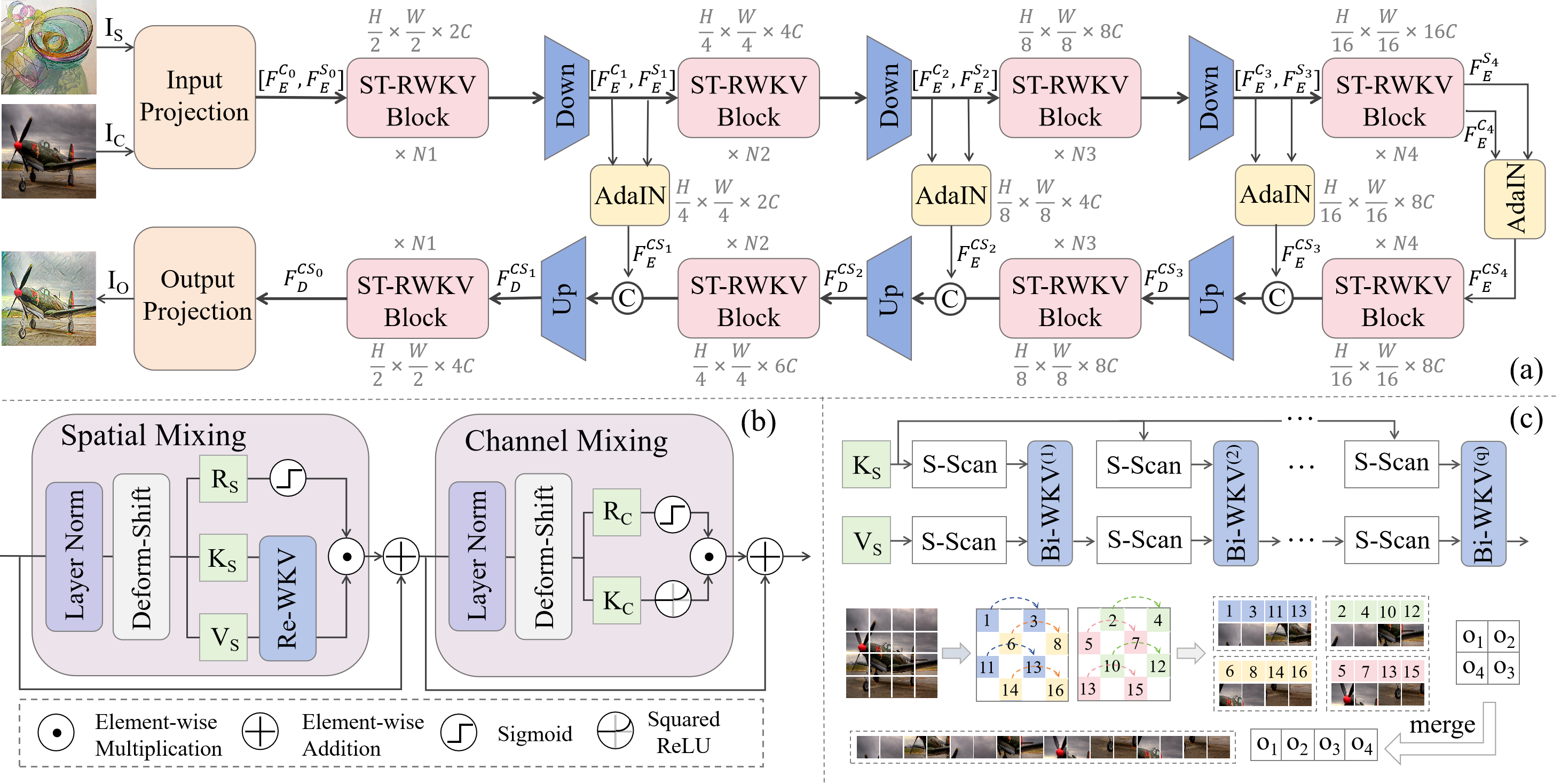}}
		\caption{(a) Style-RWKV Architecture. (b) Our ST-RWKV block incorporates a Re-WKV attention mechanism to model global dependencies with linear complexity, while a Deform-Shifting layer captures the local context within the ROIs. (c) Top: Re-WKV introduces Bi-WKV attention recurrently along the S-scanning directions, effectively achieving a global receptive field. Bottom: The S-Scanning mechanism skips samples (with a step of 2), performs an intra-group traversal, and merges inter-group sequences, enabling long-range patch connections.
		}
		\label{framework}
	\end{center}
        \vspace{-0.8cm}
\end{figure*}
\section{Related Work}
\noindent \textbf{Neural Style Transfer (NST)}~\cite{b1, b5, b6, b8, b9, b10, b11, b12, b13, b14, b15, b16, b50, b51, b52, b53, b59, b60, b61, b62, b63, b64, b65, b66, b69, b71} aims to generate a new image preserving the content but with the artistic representation of the
style source. Pioneered by Gatys et al.~\cite{b1}, a series of works have been studying this area. The mainstream methods can be categorized by the following groups: CNN-based methods~\cite{b1} aim to stylize the image with deep convolution neural networks, but they suffer from limited receptive field when the model is deeper. Then, attention-based methods~\cite{b6, b9, b11, b12, b56, b57} are introduced, and they use the self-/cross-attention mechanism to derive the semantic mapping for artistic style transfer. Nontheless, they suffer from quadratic computational complexity, which hinders its application in realistic scenarios. Recently, diffusion-based style transfer methods~\cite{b13, b14, b15, b16, b58} showed how to leverage the generative capability of diffusion models in order to perform style transfer. However, their generative process require multiple iterations for denoising, leading to inevitable time consumption.

\noindent \textbf{RWKV}~\cite{b23, b27, b28, b29, b31, b32, b34} is first proposed in NLP, it addresses memory bottleneck and quadratic scaling in Transformers through efficient linear scaling while retaining expressive characteristics like parallelized training and robust scalability. Vision-RWKV successfully transferred the RWKV from NLP to vision tasks, outperforming ViT~\cite{b26} in image classification with faster processing and reduced memory consumption for high-resolution inputs. Diffusion-RWKV~\cite{b31} adapts RWKV for DMs in image generation tasks, achieving competitive or superior performance compared to existing CNN or Transformer-based diffusion models. RWKV-CLIP~\cite{b28} achieves RWKV-driven vision-language representation learning. Meanwhile, several works~\cite{b27, b32} explore RWKV for other domains, including segmentation~\cite{b32}, point cloud learning~\cite{b27}, video understanding~\cite{b34} and more. Unfortunately, no research investigates RWKV’s potential in style transfer tasks.  To the best of our knowledge, this is the first work that studies the RWKV-like model in NST.

\section{Methodology}
The overall pipeline of StyleRWKV is shown in Fig. \ref{framework}~(a), that enables style transfer with linear complexity and limited memory usage. StyleRWKV is stacked by our presented Style Transfer RWKV (ST-RWKV) Block, including three key innovations: the Recurrent WKV attention (Re-WKV) mechanism, Deformable Shifting (Deform-Shifting), and the Skip Scanning (S-Scanning) strategy. In this work, content images, style images and stylized results are denoted as $I_C$, $I_S$ and $I_O$, respectively. $F_E^{*_i} (*\in\{C, S\}, i=0\sim4)$ represents the encoded embeddings at various stages of the encoding process, while $F_E^{{CS}_j} (j=1 \sim 4)$ refers to the embeddings generated using adaptive instance normalization (AdaIN)~\cite{b5} at different stages. Additionally, $F_D^{{CS}_k} (k=0\sim3)$ indicates the embeddings at different stages of the decoding process. 
\subsection{Hierarchical Architecture}
Our proposed Style-RWKV is a 4-level U-shaped encoder-decoder architecture with 3 times downsampling and upsampling, which enjoys the advantage of capturing image features at different hierarchies and computational efficiency. Given ${I_S}\in \mathbb{R}^{H\times W\times3}$ and ${I_C}\in \mathbb{R}^{H\times W\times3}$, we first employ a $3\times3$ convolution layer as an input projection to project images to a shallow features $F_E^{S_0}\in \mathbb{R}^{\frac{H}{2}\times \frac{W}{2}\times C}$ and $F_E^{C_0}\in \mathbb{R}^{\frac{H}{2}\times \frac{W}{2}\times C}$. Then $[F_E^{C_0}, F_E^{S_0}]$ undergoes a 4-level hierarchical encoder-decoder and is transformed into a deep feature $F_D^{{CS}_0} \in \mathbb{R}^{\frac{H}{2}\times \frac{W}{2}\times 4C}$. Each level of the encoder-decoder comprises $N_i, i\in \{1, 2, 3, 4\}$, style transfer RWKV (ST-RWKV) blocks for feature extraction. For feature downsampling, we use a $1 \times 1$ convolution layer and a pixel-unshuffle operation, reducing the spatial size by half and doubling the channel number. For upsampling, we employs a pixel-shuffle operation and a $1 \times 1$ convolution layer, doubling the spatial size and halving the channel number. In order to take full advantage of features in both shallow and deep levels, we adopt AdaIN layer that aligns the mean and variance of $F_E^{C_i}, i\in \{1, 2, 3, 4\}$ with those of $F_E^{S_i}, i\in \{1, 2, 3, 4\}$ to get $F_E^{{CS}_i}, i\in \{1, 2, 3, 4\}$ and $F_E^{{CS}_i}, i\in \{1, 2, 3\}$ are concatenated with decoder features via skip connections. Finally, $3\times3$ convolution is applied to generate the final stylized image by inverting $F_D^{{CS}_0}$ to the image space.
\subsection{Style Transfer RWKV (ST-RWKV) Block}
ST-RWKV blocks play a crucial role in extracting features at different hierarchical levels within Style-RWKV. As illustrated in Fig.~\ref{framework}~(b),  ST-RWKV incorporates both a spatial mix and a channel mix module, enabling token interaction across spatial dimensions and feature fusion across channels, respectively. Firstly, the spatial mix module plays the role of a global attention. Specifically, the input feature $F_m^n \in \mathbb{R}^{ T \times C}, m\in \{E, D\}, n \in \{C, S, CS\} $, which are first shifted and fed into 3 parallel linear layers to obtain the matrices.
\begin{equation}
\left\{
\begin{aligned}
    {R_{m}^n}_s = &\; \text{Deform-Shifting}_R(F_m^n) W_{R_s} \\
    {K_{m}^n}_s = &\; \text{Deform-Shifting}_K(F_m^n) W_{K_s} \\
    {V_{m}^n}_s = &\; \text{Deform-Shifting}_V(F_m^n) W_{V_s}
\end{aligned}
\right.
\end{equation}
were ${K_{m}^n}_s$ and ${V_{m}^n}_s$ are utilized to acquire the global attention result ${wkv_{m}^n} \in \mathbb{R}^{T \times C}$ by our proposed Recurrent Attention (Re-WKV) mechanism and multiplied with $\sigma({R_{m}^n}_s)$ which modulates the received probability:
\begin{equation}
\begin{aligned}
    {O_{m}^n}_s = &\; (\sigma({R_{m}^n}_s) \odot wkv_{m}^n) W_{O_s} \\ 
    where \; wkv_{m}^n = &\; \text{Re-WKV}({K_{m}^n}_s, {V_{m}^n}_s)
\end{aligned}
\end{equation}
where $\sigma(\cdot)$ represents the sigmoid gating, and $\bigodot$ means an element-wise multiplication. Besides, Deformable Shifting (Deform-Shifting) is a token shifting function. After an output linear projection, features are then stabilized using layer normalization. Subsequently, the tokens are passed into the channel-mix module for a channel-wise fusion. ${R_{m}^n}_c$, ${K_{m}^n}_c$ are obtained in a similar manner as spatial-mix.
\begin{equation}
\left\{
\begin{aligned}
    {R_{m}^n}_c = &\; \text{Deform-Shifting}_R({O_{m}^n}_s) W_{R_c} \\
    {K_{m}^n}_c = &\; \text{Deform-Shifting}_K({O_{m}^n}_s) W_{K_c}
\end{aligned}
\right.  
\end{equation}
where ${V_{m}^n}_c$ is estimated from ${K_{m}^n}_c$ after the activation function, and the output ${O_{m}^n}_c$ is also controlled by $\sigma({R_{m}^n}_c)$ before the output projection:
\begin{equation}
\left\{
\begin{aligned}
    {O_{m}^n}_c = &\;(\sigma({R_{m}^n}_c) \odot {V_{m}^n}_c) W_{O_c} \\ {V_{m}^n}_c = &\;\gamma({K_{m}^n}_c) W_{V_c}
    \label{eq:output_projection}
\end{aligned}
\right.  
\end{equation}
where $W_{R_c}$, $W_{K_c}$, and $W_{V_c}$ represent the three linear projection layers. $\gamma(\cdot)$ denotes the squared ReLU activation function, known for its enhanced nonlinearity.

\subsection{Recurrent Attention (Re-WKV)}
Directly applying Bidirectional WKV (Bi-WKV) attention~\cite{b29} to 2D images is challenging due to its direction-sensitive nature. According to Eq.~\ref{eq:output_projection}, Bi-WKV is partially determined by the relative position bias between tokens, indicating that Bi-WKV is sensitive to the arrangement order of sequential tokens. To overcome this limitation, as shown in the upper part of Fig.~\ref{framework}~(c), we propose a Recurrent WKV attention (Re-WKV) that introduces Bi-WKV attention along Skip Scanning (S-Scanning) directions. The attention calculation for the $t$-th token is given by the formula (5), and $\text{Bi-WKV}^{(j)}(\cdot)$ denotes the $j$-th Bi-WKV attention. $T$ denotes the total number of tokens, $w$ and $u$ are two C-dimensional learnable vectors that represent channel-wise spatial decay and the bonus indicating the current token, respectively. $k_t$ and $v_t$ represent the $t$-th features of ${K_{m}^n}_s$ and ${V_{m}^n}_s$, respectively.
\begin{equation}
\begin{aligned}
    {wkv_m^n}^{(j)} = & \; \text{Bi-WKV}^{(j)}({K_{m}^n}_s, {wkv_m^n}^{(j-1)})\\
    {wkv_m^n} = & \; \text{Bi-WKV}({K_{m}^n}_s, {V_{m}^n}_s) \\
    = & \; \frac{\sum\limits_{i=1, i \neq t}^T e^{-(|t-i|-1)/T \cdot w + k_i} v_i + e^{u + k_t} v_t}{
    \sum\limits_{i=1, i \neq t}^T e^{-(|t-i|-1)/T \cdot w + k_i} + e^{u + k_t}}
    \label{re}
\end{aligned}
\end{equation}

The final attention result $wkv$ is obtained after applying Bi-WKV attention recurrently $q$ times:
\begin{equation}
    wkv_m^n = \text{Re-WKV}({K_m^n}_s, {V_m^n}_s) = {wkv_m^n}^{(q)}
\end{equation}
where $q \ll T$, the computational complexity of Re-WKV maintains the linear computational complexity.
\subsection{Scanning and Shifting Mechanism}
\noindent {\textbf{Skip Scanning (S-Scanning).}}
A global receptive field is essential for our model to effectively capture the spatial structure within images. Conventional scanning methods, such as bi-directional~\cite{b36} and zigzag scanning~\cite{b37}, operate along fixed paths, restricting each patch to form contextual relationships solely with neighboring patches along the path. This limitation constrains the patches' receptive fields, reducing the ability to establish comprehensive spatial dependencies.

Previous research~\cite{b38, b39} highlights that atrous-based strategies can effectively expand the receptive field without sacrificing resolution. Inspired by this, we introduce the Skip Scanning (S-Scanning) method, as illustrated in the bottom of Fig.~\ref{framework}~(c). For ${K_m^n}_s \in \mathbb{R}^{C \times H \times W}, m\in \{E, D\}, n \in \{C, S, CS\} $, we scan patches with a step size $p$ and partition into selected spatial dimensional features $\{ O_i \}_{i=1}^{4}$:
\begin{equation}
\begin{aligned}
O_i \xleftarrow{\text{scan}} {K_m^n}_s[:, a :: p, \, b :: p], \quad {{K_m^n}_s}^{\prime} \xleftarrow{\text{merge}} {O}_i
\label{s-scan equation}
\end{aligned}
\end{equation}
\begin{equation}
\begin{aligned}
(a, b) = \left(
\frac{1}{2} + \frac{1}{2} \sin  \left( \frac{\pi}{2} (i-2) \right),
\frac{1}{2} + \frac{1}{2} \cos \left( \frac{\pi}{2} (i-2) \right)
\right)
\label{s-scan equation}
\end{aligned}
\end{equation}
where $O_i \in \mathbb{R}^{C \times \frac{H}{p} \times \frac{W}{p}}$ and the operation $[:, a :: p, b :: p]$ represents slicing the matrix for each channel, starting at $a$ on $H$ and $b$ on $W$, skipping every $p$ steps. Eq.~\ref{s-scan equation} is also applicable to ${V_m^n}_s$. This scanning method facilitates broader contextual relationship building, capturing richer global information while preserving local detail.
\begin{figure}[htbp]
	\begin{center}
		\centerline{\includegraphics[width=80mm]{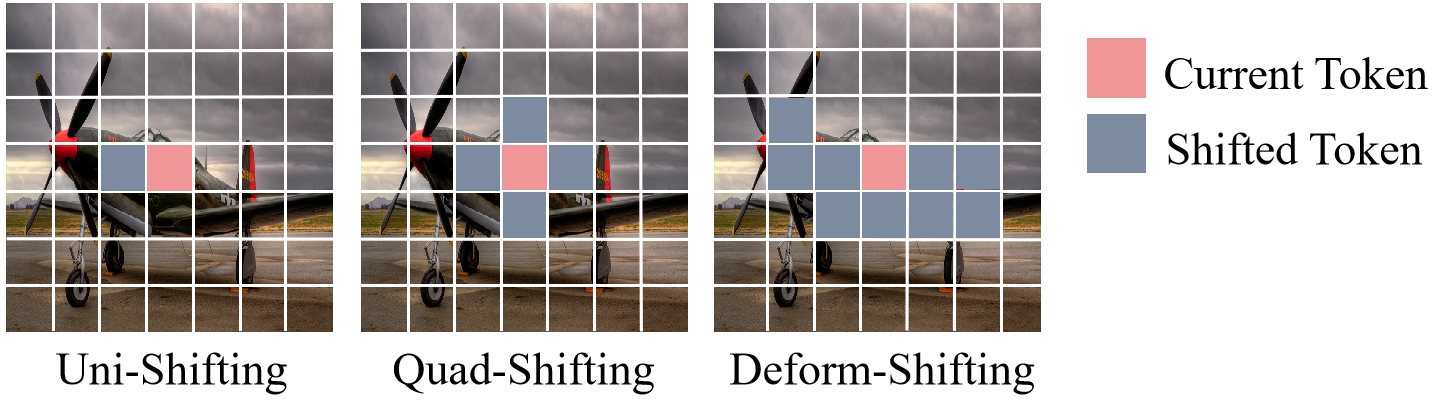}}
		\caption{
      Illustrations of different token shifting mechanisms.}
		\label{deform-shift_figure}
	\end{center}
    \vspace{-5mm}
\end{figure}

\noindent {\textbf{Deformable Shifting (Deform-Shifting).}}
 Existing token shifting mechanisms, such as the Unidirectional Shifting (Uni-Shifting)~\cite{b23} and the Quad-directional Shifting (Quad-Shifting)~\cite{b29}, as shown in Fig. \ref{deform-shift_figure}, are proposed to capture the local context in a token sequence, where neighboring tokens are assumed to be correlated and share similar context information. They only shift tokens from limited directions and do not fully exploit the spatial relationships.
 \begin{figure*}[ht]
\centering
\includegraphics[width=\linewidth]{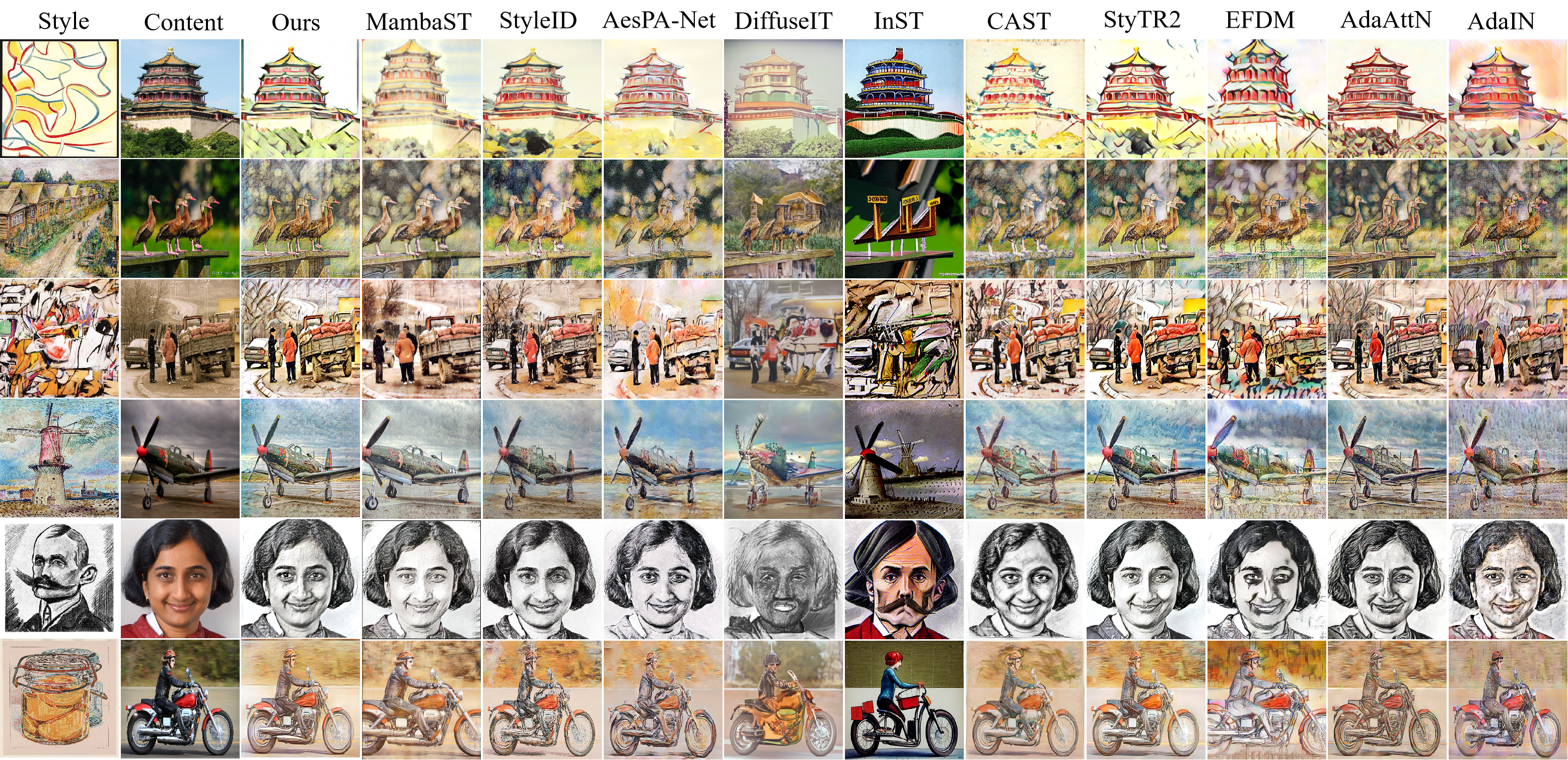}
\caption{Qualitative comparison with state-of-the-art neural style transfer methods, \emph{e.g.,} Feedforward Neural Network (FFN)- (AesPA-Net, CAST, EFDM, AdaAttN, AdaIN), DMs- (StyleID, DiffuseIT, InST), Transformer-based (StyTR$^2$), and Mamba-based (MambaST) methods.}
\label{Qualitative comparison}
\end{figure*}
 To address this issue, we introduce Deformable Shifting (Deform-Shifting) mechanism that shifts and fuses neighboring tokens from region of interest using deformable convolution~\cite{b40} which dynamically learns offsets to adjust neighboring tokens. For each current token $p_0$, we have convolution:
 \begin{equation}
\begin{aligned}
    y(p_0) = \sum_{{p_n}\in \mathcal{R}}w(p_n)\cdot x(p_0+p_n+\Delta p_n)
\end{aligned}
\end{equation}
where $p_n$ enumerates the token locations in $\mathcal{R}$. The sampling neighboring token is on the irregular and offset locations $p_n+\Delta p_n$. 
\subsection{Network Optimization}

The generated results should maintain the original content structures and the reference style patterns. Therefore, we construct two different perceptual loss terms to measure the content difference between the output image $I_o$ and the input content image $I_c$, as well as the style difference between $I_o$ and the input style reference $I_s$.

We use feature maps extracted by a pretrained VGG model to construct the content loss and the style loss following. The content perceptual loss $L_c$ is defined as
\begin{equation}
L_c = \frac{1}{N_l} \sum_{i=0}^{N_l} \| \phi_i(I_o) - \phi_i(I_c) \|_2,
\end{equation}
where $\phi_i(\cdot)$ denotes features extracted from the $i$-th layer in a pretrained VGG19 and $N_l$ is the number of layers.

The style perceptual loss $L_s$ is defined as
\begin{align}
L_s = \frac{1}{N_l} \sum_{i=0}^{N_l} &\|\mu(\phi_i(I_o)) - \mu(\phi_i(I_s))\|_2 \notag \\
&+ \|\sigma(\phi_i(I_o)) - \sigma(\phi_i(I_s))\|_2,
\end{align}
where $\mu(\cdot)$ and $\sigma(\cdot)$ denote the mean and variance of extracted features, respectively.

We also adopt identity loss to learn richer and more accurate content and style representations. Specifically, we take two of the same content (style) images into StyleRWKV, and the generated output $I_{cc}(I_{ss})$ should be identical to the input $I_c(I_s)$. Therefore, we compute two identity loss terms to measure the differences between $I_c(I_s)$ and $I_{cc}(I_{ss})$:
\begin{align}
L_{\text{id1}} &= \| I_{cc} - I_c \|_2 + \| I_{ss} - I_s \|_2, \\
L_{\text{id2}} &= \frac{1}{N_l} \sum_{i=0}^{N_l} \| \phi_i(I_{cc}) - \phi_i(I_c) \|_2 + \| \phi_i(I_{ss}) - \phi_i(I_s) \|_2.
\end{align}

The entire network is optimized by minimizing the following function:
\begin{equation}
\mathcal{L} = \lambda_c L_c + \lambda_s L_s + \lambda_{\text{id1}} L_{\text{id1}} + \lambda_{\text{id2}} L_{\text{id2}}.
\end{equation}
We set $\lambda_c, \lambda_s, \lambda_{\text{id1}},$ and $\lambda_{\text{id2}}$ to alleviate the impact of magnitude differences.

\begin{table*}[htp]
\setlength{\belowcaptionskip}{-0.2cm}
\centering
\caption{Quantitative comparison in ArtFID, FID, LPIPS, user preference, inference time and parameter.}
\renewcommand{\arraystretch}{1.2}
\resizebox{1.0\linewidth}{!}{
\begin{tabular}{c|c|c|c|c|c|c|c|c|c|c|c|c}
\hline
\multicolumn{2}{c|}{Metrics} & {Ours} & {Mamba-ST~\cite{b47}} & {StyleID~\cite{b16}} & {AesPA-Net~\cite{b12}} & {DiffuseIT~\cite{b15}} & {InST~\cite{b13}} & {CAST~\cite{b46}} & {StyTR$^2$~\cite{b11}} & {EFDM~\cite{b10}} & {AdaAttN~\cite{b9}} & {AdaIN~\cite{b5}} \\
\hline
\multicolumn{2}{c|}{ArtFID $\downarrow$} & \textbf{26.370} & 32.178 & 33.979 & 34.865 & 87.302 & 59.418 & 41.873 & 31.607 & 33.851 & 35.889 & 44.866 \\
\multicolumn{2}{c|}{FID $\downarrow$} & \textbf{16.362} & 19.980 & 19.531 & 19.840 & 30.168 & 31.683 & 23.865 & 19.890 & 21.096 & 21.657 & 25.584 \\
\multicolumn{2}{c|}{LPIPS $\downarrow$} & \textbf{0.451} & 0.617 & 0.655 & 0.673 & 1.801 & 0.818 & 0.684 & 0.513 & 0.532 & 0.584 & 0.690 \\
\hline
\multicolumn{2}{c|}{Preference}& {0.7$^*$} & \textbf{0.66}/0.34 & \textbf{0.68}/0.32 & \textbf{0.72}/0.28 & \textbf{0.83}/0.17 & \textbf{0.89}/0.11 & \textbf{0.72}/0.28 & \textbf{0.65}/0.35 & \textbf{0.78}/0.22 & \textbf{0.71}/0.29 & \textbf{0.84}/0.16 \\
\hline
\multirow{4}{*}{Inference/sec} & 128$\times$128 & \textbf{0.266} & 3.587 & 17.153 & 0.668 & 14.262 & 10.371 & 4.149 & 10.576 & 2.044 & 2.422 &2.837\\ 
 \multirow{4}{*}{} & 256$\times$256 & \textbf{0.638} & 9.247 & 17.328 & 0.672 & 14.691 & 10.425 & 4.204 & 169.116 & 2.057 & 2.571 &2.912\\ 
\multirow{4}{*}{} & 512$\times$512 & 1.530 & 30.836 & 17.343 & \textbf{0.691} & 15.723 & 12.235 & 4.534 & 2707.345 & 2.074 & 2.808 &3.013\\ 
\multirow{4}{*}{} & 1024$\times$1024 & 3.670 & 120.344 & 17.416 & \textbf{0.695} & 17.870 & 15.482 & 5.875 & 43319.256 & 2.169 & 4.304 &3.142\\ \hline
\multicolumn{2}{c|}{ Parameter/M}& {28.80} & {40.03} & {1066.25} & {11.25} & {552.81} & {2602.69} & {10.52} & {35.39} & \textbf{3.51} & {13.63} & {7.01} \\
\hline
\multicolumn{12}{l}{Note: * represents the average user responses. 128$\times$128, 256$\times$256, etc. refer to sequence lengths rather than resolution.}
\end{tabular}}
\label{FID}
\vspace{-0.3cm}
\end{table*}

\section{Experiments}
\subsection{Experimental Setting}\label{experimental_setting}
\noindent {\textbf{Implementation Details.}} 
In our experiments, the number of ST-RWKV blocks is set to N$_1$ = 4, N$_2$ = N$_3$ = 6, and N$_4$ = 8. The number of input channels is C = 48, and the recurrence number of Re-WKV is q = 2. We use 70,097 content images from MS-COCO~\cite{b41} and 70,097 style images from WikiArt~\cite{b42} as the training datasets. The training process is conducted 150k steps with a batch size of 2. We employ the Adam optimizer with a learning rate of \emph{1e-4} and resize the input images to 512 $\times$ 512 during training. We set $\lambda_c, \lambda_s, \lambda_{\text{id1}},$ and $\lambda_{\text{id2}}$ to $8, 15, 100,$ and $1$ to alleviate the impact of magnitude differences. Additionally, we use 5,000 pairs of COCO-WikiArt images for testing. 

\noindent {\textbf{Evaluation Metrics.}} 
We employ three metrics to evaluate model performance: 
(1) LPIPS~\cite{b43} measures content fidelity between the stylized image and the corresponding content image. 
(2) FID~\cite{b44} assesses the style fidelity between the stylized image and the corresponding style image. (3) ArtFID~\cite{b45} which evaluates overall style transfer performances with consideration of both content and style preservation and also is known as strongly coinciding with human judgment. Specifically, ArtFID is computed as:
\begin{equation}
ArtFID = (1 + LPIPS) \cdot (1 + FID)
\end{equation}
\subsection{Comparison Results}

\noindent \textbf{Qualitative Comparison.}

- \emph{Comparison with state-of-the-art neural style transfer methods.}
As shown in the Fig. \ref{Qualitative comparison}, our method effectively preserves the structure and details of the content image, while ensuring overall consistency in style transfer due to the global receptive field. In comparison, other methods often lose the structure of content or fail to transfer the style. Specifically, DiffuseIT suffers from generating shape and visually plausible images. InST drops the original content and struggles to transfer style. In the 2nd row, our stylized results effectively preserve the red beak and red legs of the duck in the content image, demonstrating the effectiveness of Deform-Shifting in modeling local spatial relationships in the region of interest. 
Comparatively, other methods either alter the color of the beak or legs, or exhibit blurred edges and poor detail preservation. Furthermore, in the 3rd row, our method maintains clear edges for each object with good fidelity, demonstrating the excellent sharpening capability of Re-WKV, while ensuring that the textures and colors align well with the style image. Nevertheless, other methods such as StyTR$^2$ and EFDM introduce undesired green artifacts, while AdaAttN causes an over-sharpening effect, resulting in distorted relative spatial relationships between objects.

- \emph{Comparisons of Style, Content, and Stylization Result.}
Fig. 5 presents a vertical visual comparison. The first three columns show a comparison of results generated from the same content image with different style images, while the last three columns compare the stylized results of the same style image applied to different content images. 
From the first three columns, it can be seen that our method effectively preserves the content structure and details across various styles, with a clear emphasis on the main subject and a blurred background, resulting in high-quality stylization. From the last three columns, it is evident that our stylized results exhibit good consistency in color and texture transfer across different content images.

\noindent \textbf{Quantitative Comparison.}

- \emph{Stylization quality}.
As shown in rows 2nd-4th of the Tab.~\ref{FID}, our method demonstrates the best performance in terms of LPIPS, FID, and ArtFID. Specifically, our method largely surpasses other methods in terms of ArtFID, which is known as coinciding the human preference. Additionally, the proposed method records the lowest FID, which denotes that stylized images highly resemble the target styles. For content fidelity metrics, ours shows superior scores in LPIPS. 

\begin{table}[htbp!]
\centering
\caption{Quantative ablations on different recurrence numbers.}
\resizebox{8cm}{!}{
\begin{tabular}{c|c|c|c|c}
\hline
\textbf{Recurrence Number} & \textbf{ArtFID ↓} & \textbf{FID ↓} & \textbf{LPIPS ↓} & \textbf{Time/s} \\ \hline
$q=1$ & 27.783 & 18.128 & 0.593 & 0.205 \\ 
$q=2$ (ours) & 26.370 & 16.362 & 0.451 & 0.213 \\ 
$q=3$ & 25.639 & 15.442 & 0.448 & 0.236 \\  \hline
\end{tabular}}
\label{table:recurrence}
\end{table}
\begin{figure}[ht]
		\centerline{\includegraphics[width=80mm]{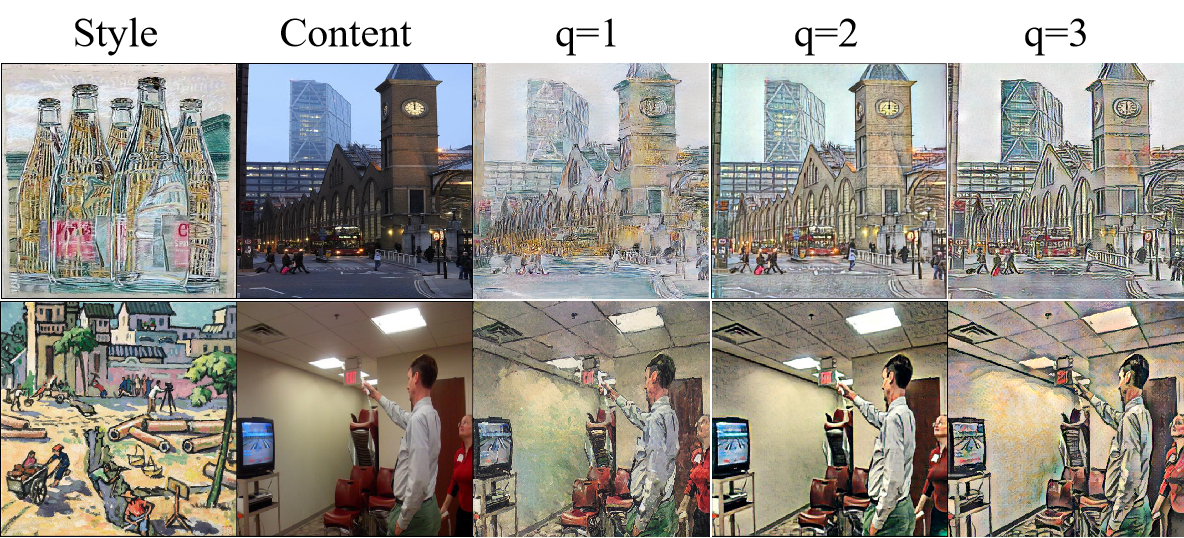}}
		\caption{
       Qualitative ablations on different recurrence number.}
		\label{fig_recurrence}
\end{figure}

\begin{figure}[ht]
  \centering  
  \includegraphics[width=\linewidth]{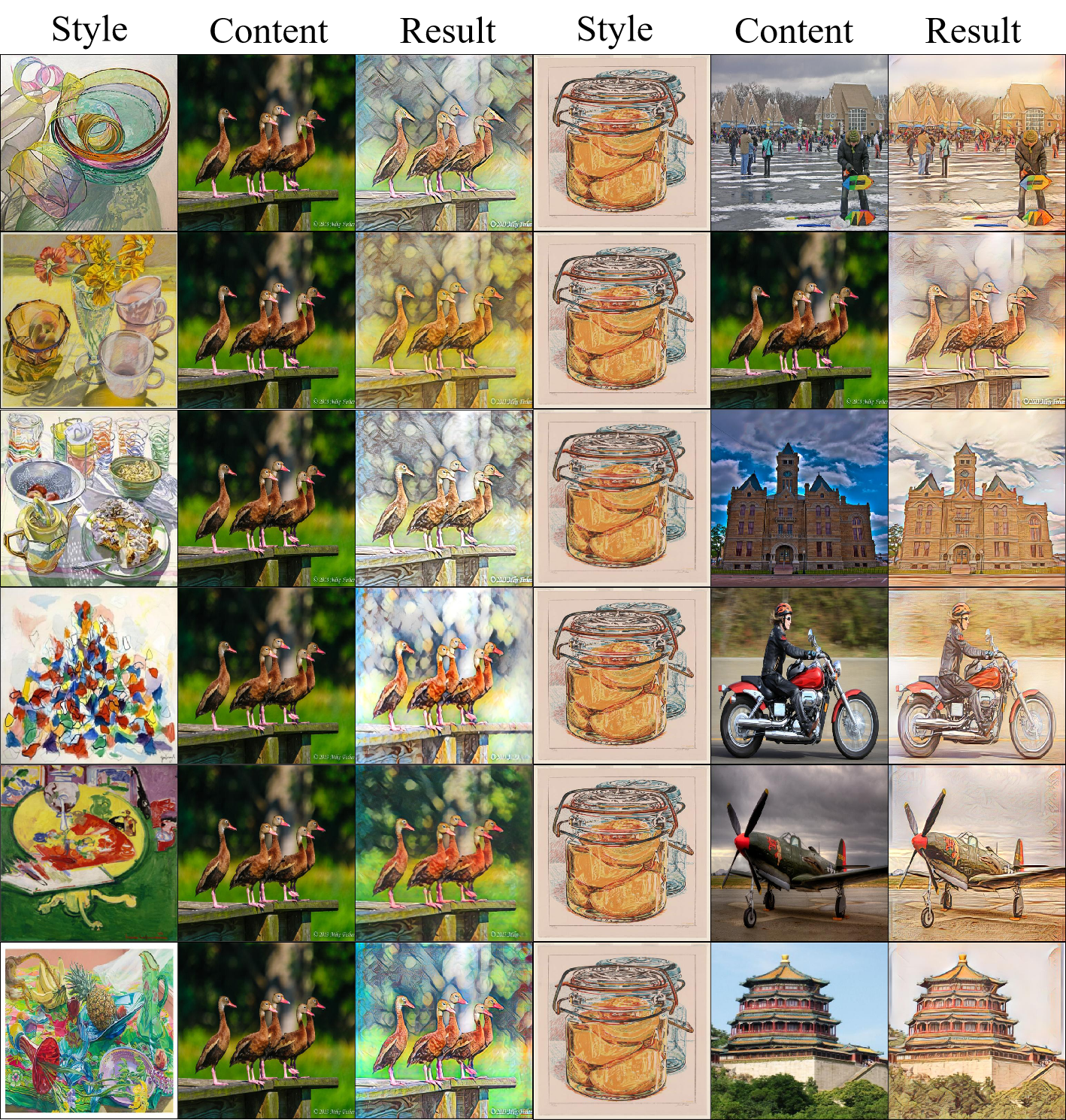}
  \caption{Comparisons of Style, Content, and Stylization Result.}
  \label{supp-fig2}
\end{figure}

- \emph{User preference}. We conducted a user study to further validate our method. We sampled 45 stylized results from random pairs of content and style images, and conduct one-to-one comparisons with ten sota methods. And we collected 2250 responses from 50 subjects. As illustrated in the 5th row of the Tab.~\ref{FID}, most participants favored our stylized results over the state-of-the-art competitors.

- \emph{Efficiency and Model complexity}. As seen in rows 6th-9th of the Tab.~\ref{FID}. FFN-based methods achieve the highest inference efficiency with the lowest model complexity. Diffusion-based methods have the longest inference time and a larger number of model parameters. Transformer-based method exhibits quadratic complexity \emph{w.r.t} sequence length. Both Mamba and RWKV have linear complexity, but in long-sequence modeling (high-resolution images), our method demonstrates higher inference efficiency than MambaST.

\begin{table}[ht]
\centering
\caption{Comparison of different token shifting mechanisms.}
\resizebox{8cm}{!}{
\begin{tabular}{c|c|c|c|c}
\hline
\textbf{Shifting Mechanism} & \textbf{ArtFID ↓} & \textbf{FID ↓} & \textbf{LPIPS ↓} & \textbf{Time/s} \\ \hline 
Quad-Shifting & 29.101 & 20.575 & 0.769 & 0.201  \\ 
Omni-Shifting & 26.781 & 17.258 & 0.484 & 0.205 \\
Deform-Shifting (Ours) & 26.370 & 16.362 & 0.451 & 0.213 \\
\hline
\end{tabular}}
\label{table_shift}
\end{table}

\begin{figure}[h!]
  \centering  
  \includegraphics[width=\linewidth]{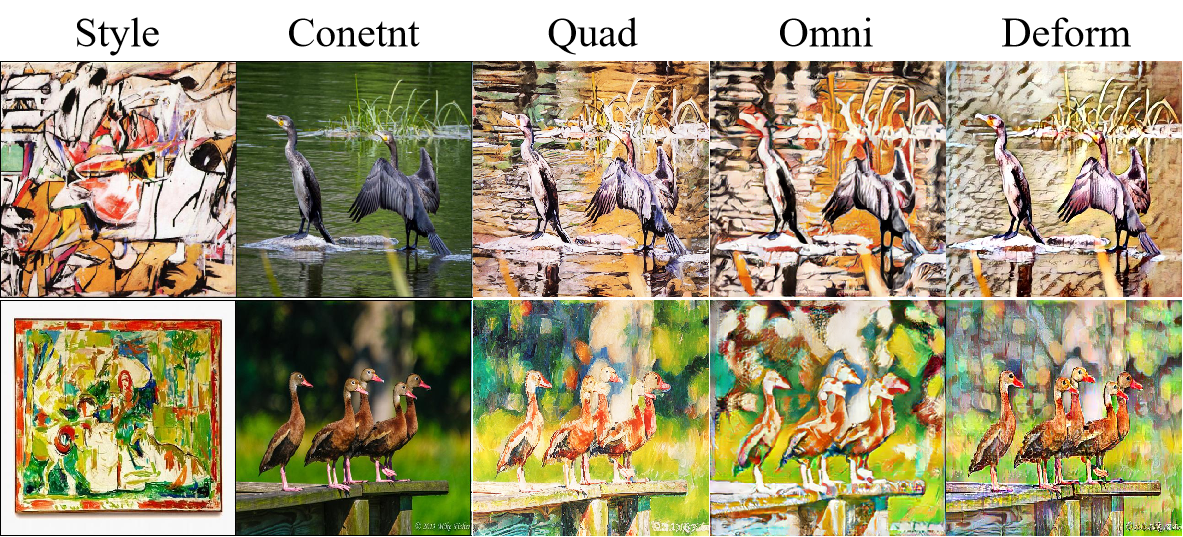}
  \caption{Ablations on different token shifting methods.} 
  \label{fig_shfit}
\end{figure}

\begin{table}[]
\centering
\caption{Comparison of different scanning mechanisms.}
\resizebox{8cm}{!}{
\begin{tabular}{c|c|c|c|c}
\hline
\textbf{Scanning Mechanism} & \textbf{ArtFID ↓} & \textbf{FID ↓} & \textbf{LPIPS ↓} & \textbf{Time/s} \\ \hline
BiDirectional & 28.610 & 16.811 & 0.578 & 0.336 \\
Zigzag & 27.019 & 16.901 & 0.581 & 0.468 \\ 
Skipping (ours) & 26.370 & 16.362 & 0.451 & 0.213 \\ \hline
$p=1$ & 30.838 & 17.395 & 0.693 & 0.217 \\
$p=2$ (ours) & 26.370 & 16.362 & 0.451 & 0.213 \\
$p=3$ & 27.092 & 16.381 & 0.502 & 0.243 \\ \hline
\end{tabular}}
\label{table_scan1}
\vspace{-0.4cm}
\end{table}

\subsection{Ablation Study}
\noindent \textbf{Impact of Recurrence number.}
Tab. ~\ref{table:recurrence} and Fig. \ref{fig_recurrence} show quantitative and qualitative experiments with recurrence numbers (\textit{q}) of 1, 2, and 3. 
The results demonstrate that increasing \textit{q} of Re-WKV enhances the model's global modeling capabilities. This is reflected in the decreased ArtFID, FID, and LPIPS metrics, the edges of objects in stylization results become increasingly clearer. However, the inference time also increases with the number of iterations. Consequently, we selected \textit{q}=2 which provides an optimal trade-off between performance and inference efficiency.

\noindent \textbf{Effectiveness of Deform-Shifting.}
Tab.~\ref{table_shift} and Fig.~\ref{fig_shfit} show the quantitative and qualitative comparisons of Quad-Shifting, Omni-Shifting, and Deform-Shifting. The results demonstrate that Deform-Shifting achieves the better performance, indicating its effectiveness in emphasizing and enhancing the stylization of RoIs. However, compared to the other ones, it slightly increases the inference time.

\noindent \textbf{S-Scanning Analysis.}
We conducted a quantitative study on the scanning method and its performance with varying skip steps, as shown in Table~\ref{table_scan1}. The S-Scanning method can establish spatial relationships between distant patches, allowing it to integrate information from different regions and thereby improve the consistency of the style transfer. Thus, our scanning method surpasses both bidirectional and zigzag scanning methods. Furthermore, when the skip step p=2, it strikes a good balance between performance and efficiency.

\section{Conclusion}
We propose a novel style transfer method, StyleRWKV, using a hierarchical RWKV-like architecture. Our approach explores three key innovations: the Re-WKV attention mechanism, which captures global dependencies with linear complexity; Deform-Shifting, which dynamically shifts tokens to better capture local dependencies within an image’s object structures; and the S-Scanning mechanism, which effectively models distant contextual dependencies. Extensive experiments confirm the effectiveness of these designs, with comparisons revealing superior stylization quality, lower model complexity, and higher inference efficiency.
{
    \small
    \bibliographystyle{ieeenat_fullname}
    \bibliography{main}

\begin{thebibliography}{60}
\providecommand{\natexlab}[1]{#1}
\providecommand{\url}[1]{\texttt{#1}}
\expandafter\ifx\csname urlstyle\endcsname\relax
  \providecommand{\doi}[1]{doi: #1}\else
  \providecommand{\doi}{doi: \begingroup \urlstyle{rm}\Url}\fi

\bibitem[An et~al.(2021)An, Huang, Song, Dou, Liu, and Luo]{b62}
Jie An, Siyu Huang, Yibing Song, Dejing Dou, Wei Liu, and Jiebo Luo.
\newblock Artflow: Unbiased image style transfer via reversible neural flows.
\newblock In \emph{Proceedings of the IEEE/CVF Conference on Computer Vision and Pattern Recognition}, pages 862--871, 2021.

\bibitem[Botti et~al.(2024{\natexlab{a}})Botti, Ergasti, Rossi, Fontanini, Ferrari, Bertozzi, and Prati]{b47}
Filippo Botti, Alex Ergasti, Leonardo Rossi, Tomaso Fontanini, Claudio Ferrari, Massimo Bertozzi, and Andrea Prati.
\newblock Mamba-st: State space model for efficient style transfer.
\newblock \emph{arXiv preprint arXiv:2409.10385}, 2024{\natexlab{a}}.

\bibitem[Botti et~al.(2024{\natexlab{b}})]{b24}
Filippo Botti et~al.
\newblock Mamba-st: State space model for efficient style transfer.
\newblock \emph{arXiv preprint arXiv:2409.10385}, 2024{\natexlab{b}}.

\bibitem[Caesar et~al.(2018)Caesar, Uijlings, and Ferrari]{b41}
Holger Caesar, Jasper Uijlings, and Vittorio Ferrari.
\newblock Coco-stuff: Thing and stuff classes in context.
\newblock In \emph{CVPR}, pages 1209--1218, 2018.

\bibitem[Chen et~al.(2021{\natexlab{a}})Chen, Wang, Zhang, Zuo, Li, Xing, Lu, et~al.]{b63}
Haibo Chen, Zhizhong Wang, Huiming Zhang, Zhiwen Zuo, Ailin Li, Wei Xing, Dongming Lu, et~al.
\newblock Artistic style transfer with internal-external learning and contrastive learning.
\newblock \emph{Advances in Neural Information Processing Systems}, 34:\penalty0 26561--26573, 2021{\natexlab{a}}.

\bibitem[Chen et~al.(2021{\natexlab{b}})Chen, Wang, Zhang, Zuo, Li, Xing, Lu, et~al.]{b64}
Haibo Chen, Zhizhong Wang, Huiming Zhang, Zhiwen Zuo, Ailin Li, Wei Xing, Dongming Lu, et~al.
\newblock Artistic style transfer with internal-external learning and contrastive learning.
\newblock \emph{Advances in Neural Information Processing Systems}, 34:\penalty0 26561--26573, 2021{\natexlab{b}}.

\bibitem[Chung et~al.(2024)Chung, Hyun, and Heo]{b16}
Jiwoo Chung, Sangeek Hyun, and Jae-Pil Heo.
\newblock Style injection in diffusion: A training-free approach for adapting large-scale diffusion models for style transfer.
\newblock In \emph{CVPR}, pages 8795--8805, 2024.

\bibitem[Dai et~al.(2017)Dai, Qi, Xiong, Li, Zhang, Hu, and Wei]{b40}
Jifeng Dai, Haozhi Qi, Yuwen Xiong, Yi Li, Guodong Zhang, Han Hu, and Yichen Wei.
\newblock Deformable convolutional networks.
\newblock In \emph{ICCV}, pages 764--773, 2017.

\bibitem[Dai et~al.(2025)Dai, Zhou, Yi, and Ma]{b71}
Miaomiao Dai, Qianyu Zhou, Ran Yi, and Lizhuang Ma.
\newblock Diffusefist: A fast image-guided style transfer method for adapting large-scale diffusion models.
\newblock In \emph{IEEE International Conference on Acoustics, Speech and Signal Processing (ICASSP)}, 2025.

\bibitem[Dai et~al.(2019)Dai, Liang, Qiu, and Huang]{b54}
Ning Dai, Jianze Liang, Xipeng Qiu, and Xuanjing Huang.
\newblock Style transformer: Unpaired text style transfer without disentangled latent representation.
\newblock \emph{arXiv preprint arXiv:1905.05621}, 2019.

\bibitem[Deng et~al.(2020)Deng, Tang, Dong, Sun, Huang, and Xu]{b60}
Yingying Deng, Fan Tang, Weiming Dong, Wen Sun, Feiyue Huang, and Changsheng Xu.
\newblock Arbitrary style transfer via multi-adaptation network.
\newblock In \emph{Proceedings of the 28th ACM international conference on multimedia}, pages 2719--2727, 2020.

\bibitem[Deng et~al.(2021)Deng, Tang, Dong, Huang, Ma, and Xu]{b61}
Yingying Deng, Fan Tang, Weiming Dong, Haibin Huang, Chongyang Ma, and Changsheng Xu.
\newblock Arbitrary video style transfer via multi-channel correlation.
\newblock In \emph{Proceedings of the AAAI Conference on Artificial Intelligence}, pages 1210--1217, 2021.

\bibitem[Deng et~al.(2022)Deng, Tang, Dong, Ma, Pan, Wang, and Xu]{b11}
Yingying Deng, Fan Tang, Weiming Dong, Chongyang Ma, Xingjia Pan, Lei Wang, and Changsheng Xu.
\newblock Stytr2: Image style transfer with transformers.
\newblock In \emph{CVPR}, pages 11326--11336, 2022.

\bibitem[Dosovitskiy(2020)]{b26}
Alexey Dosovitskiy.
\newblock An image is worth 16x16 words: Transformers for image recognition at scale.
\newblock \emph{arXiv preprint arXiv:2010.11929}, 2020.

\bibitem[Duan et~al.(2024)Duan, Wang, Chen, Zhu, Lu, Lu, Qiao, Li, Dai, and Wang]{b29}
Yuchen Duan, Weiyun Wang, Zhe Chen, Xizhou Zhu, Lewei Lu, Tong Lu, Yu Qiao, Hongsheng Li, Jifeng Dai, and Wenhai Wang.
\newblock Vision-rwkv: Efficient and scalable visual perception with rwkv-like architectures.
\newblock \emph{arXiv preprint arXiv:2403.02308}, 2024.

\bibitem[Fei et~al.(2024)Fei, Fan, Yu, Li, and Huang]{b31}
Zhengcong Fei, Mingyuan Fan, Changqian Yu, Debang Li, and Junshi Huang.
\newblock Diffusion-rwkv: Scaling rwkv-like architectures for diffusion models.
\newblock \emph{arXiv preprint arXiv:2404.04478}, 2024.

\bibitem[Gatys et~al.(2016)Gatys, Ecker, and Bethge]{b1}
Leon~A Gatys, Alexander~S Ecker, and Matthias Bethge.
\newblock Image style transfer using convolutional neural networks.
\newblock In \emph{CVPR}, pages 2414--2423, 2016.

\bibitem[Gu and Dao(2023)]{b22}
Albert Gu and Tri Dao.
\newblock Mamba: Linear-time sequence modeling with selective state spaces.
\newblock \emph{arXiv preprint arXiv:2312.00752}, 2023.

\bibitem[Gu et~al.(2024)]{b28}
Tiancheng Gu et~al.
\newblock Rwkv-clip: A robust vision-language representation learner.
\newblock \emph{arXiv preprint arXiv:2406.06973}, 2024.

\bibitem[Guo et~al.(2021)Guo, Zhou, Zhou, Gu, Tang, Feng, and Ma]{b69}
Shaohua Guo, Qianyu Zhou, Ye Zhou, Qiqi Gu, Junshu Tang, Zhengyang Feng, and Lizhuang Ma.
\newblock Label-free regional consistency for image-to-image translation.
\newblock In \emph{IEEE International Conference on Multimedia and Expo}, pages 1--6. IEEE, 2021.

\bibitem[He et~al.(2024)]{b27}
Qingdong He et~al.
\newblock Pointrwkv: Efficient rwkv-like model for hierarchical point cloud learning.
\newblock \emph{arXiv preprint arXiv:2405.15214}, 2024.

\bibitem[Heusel et~al.(2017)Heusel, Ramsauer, Unterthiner, Nessler, and Hochreiter]{b44}
Martin Heusel, Hubert Ramsauer, Thomas Unterthiner, Bernhard Nessler, and Sepp Hochreiter.
\newblock Gans trained by a two time-scale update rule converge to a local nash equilibrium.
\newblock \emph{NeurIPS}, 30, 2017.

\bibitem[Ho et~al.(2020)Ho, Jain, and Abbeel]{b19}
Jonathan Ho, Ajay Jain, and Pieter Abbeel.
\newblock Denoising diffusion probabilistic models.
\newblock \emph{NeurIPS}, 33:\penalty0 6840--6851, 2020.

\bibitem[Hong et~al.(2023)]{b12}
Kibeom Hong et~al.
\newblock Aespa-net: Aesthetic pattern-aware style transfer networks.
\newblock In \emph{ICCV}, pages 22758--22767, 2023.

\bibitem[Hu et~al.(2024)Hu, Baumann, Gui, Grebenkova, Ma, Fischer, and Ommer]{b37}
Vincent~Tao Hu, Stefan~Andreas Baumann, Ming Gui, Olga Grebenkova, Pingchuan Ma, Johannes Fischer, and Bjorn Ommer.
\newblock Zigma: Zigzag mamba diffusion model.
\newblock \emph{arXiv preprint arXiv:2403.13802}, 2024.

\bibitem[Huang and Belongie(2017)]{b5}
Xun Huang and Serge Belongie.
\newblock Arbitrary style transfer in real-time with adaptive instance normalization.
\newblock In \emph{ICCV}, pages 1501--1510, 2017.

\bibitem[Jeong et~al.(2023)Jeong, Kwon, and Uh]{b58}
Jaeseok Jeong, Mingi Kwon, and Youngjung Uh.
\newblock Training-free style transfer emerges from h-space in diffusion models.
\newblock \emph{arXiv preprint arXiv:2303.15403}, 3, 2023.

\bibitem[Johnson et~al.(2016)Johnson, Alahi, and Fei-Fei]{b51}
Justin Johnson, Alexandre Alahi, and Li Fei-Fei.
\newblock Perceptual losses for real-time style transfer and super-resolution.
\newblock In \emph{Computer Vision--ECCV 2016: 14th European Conference, Amsterdam, The Netherlands, October 11-14, 2016, Proceedings, Part II 14}, pages 694--711. Springer, 2016.

\bibitem[Kolkin et~al.(2019)Kolkin, Salavon, and Shakhnarovich]{b65}
Nicholas Kolkin, Jason Salavon, and Gregory Shakhnarovich.
\newblock Style transfer by relaxed optimal transport and self-similarity.
\newblock In \emph{Proceedings of the IEEE/CVF conference on computer vision and pattern recognition}, pages 10051--10060, 2019.

\bibitem[Kwon and Ye(2022)]{b15}
Gihyun Kwon and Jong~Chul Ye.
\newblock Diffusion-based image translation using disentangled style and content representation.
\newblock \emph{arXiv preprint arXiv:2209.15264}, 2022.

\bibitem[Li and Wand(2016)]{b50}
Chuan Li and Michael Wand.
\newblock Combining markov random fields and convolutional neural networks for image synthesis.
\newblock In \emph{Proceedings of the IEEE conference on computer vision and pattern recognition}, pages 2479--2486, 2016.

\bibitem[Li et~al.(2019)Li, Liu, Kautz, and Yang]{b66}
Xueting Li, Sifei Liu, Jan Kautz, and Ming-Hsuan Yang.
\newblock Learning linear transformations for fast image and video style transfer.
\newblock In \emph{Proceedings of the IEEE/CVF Conference on Computer Vision and Pattern Recognition}, pages 3809--3817, 2019.

\bibitem[Li et~al.(2017)Li, Fang, Yang, Wang, Lu, and Yang]{b8}
Yijun Li, Chen Fang, Jimei Yang, Zhaowen Wang, Xin Lu, and Ming-Hsuan Yang.
\newblock Universal style transfer via feature transforms.
\newblock \emph{NeurIPS}, 30, 2017.

\bibitem[Liu et~al.(2021)]{b9}
Songhua Liu et~al.
\newblock Adaattn: Revisit attention mechanism in arbitrary neural style transfer.
\newblock In \emph{ICCV}, pages 6649--6658, 2021.

\bibitem[Long et~al.(2024)Long, Zhou, Li, Lu, Ying, Luo, Ma, and Yan]{b67}
Shaocong Long, Qianyu Zhou, Xiangtai Li, Xuequan Lu, Chenhao Ying, Yuan Luo, Lizhuang Ma, and Shuicheng Yan.
\newblock Dgmamba: Domain generalization via generalized state space model.
\newblock In \emph{Proceedings of the 32nd ACM International Conference on Multimedia}, pages 3607--3616, 2024.

\bibitem[Park and Lee(2019)]{b57}
Dae~Young Park and Kwang~Hee Lee.
\newblock Arbitrary style transfer with style-attentional networks.
\newblock In \emph{proceedings of the IEEE/CVF conference on computer vision and pattern recognition}, pages 5880--5888, 2019.

\bibitem[Peng et~al.(2023)Peng, Alcaide, et~al.]{b23}
Bo Peng, Alcaide, et~al.
\newblock Rwkv: Reinventing rnns for the transformer era.
\newblock \emph{arXiv preprint arXiv:2305.13048}, 2023.

\bibitem[Rombach et~al.(2022)Rombach, Blattmann, Lorenz, Esser, and Ommer]{b21}
Robin Rombach, Andreas Blattmann, Dominik Lorenz, Patrick Esser, and Bj{\"o}rn Ommer.
\newblock High-resolution image synthesis with latent diffusion models.
\newblock In \emph{CVPR}, pages 10684--10695, 2022.

\bibitem[Saleh and Elgammal(2015)]{b42}
Babak Saleh and Ahmed Elgammal.
\newblock Large-scale classification of fine-art paintings: Learning the right metric on the right feature.
\newblock \emph{arXiv preprint arXiv:1505.00855}, 2015.

\bibitem[Sheng et~al.(2018)Sheng, Lin, Shao, and Wang]{b59}
Lu Sheng, Ziyi Lin, Jing Shao, and Xiaogang Wang.
\newblock Avatar-net: Multi-scale zero-shot style transfer by feature decoration.
\newblock In \emph{Proceedings of the IEEE conference on computer vision and pattern recognition}, pages 8242--8250, 2018.

\bibitem[Song et~al.(2020)Song, Meng, and Ermon]{b53}
Jiaming Song, Chenlin Meng, and Stefano Ermon.
\newblock Denoising diffusion implicit models.
\newblock \emph{arXiv preprint arXiv:2010.02502}, 2020.

\bibitem[Tu et~al.(2022)Tu, Talebi, Zhang, Yang, Milanfar, Bovik, and Li]{b38}
Zhengzhong Tu, Hossein Talebi, Han Zhang, Feng Yang, Peyman Milanfar, Alan Bovik, and Yinxiao Li.
\newblock Maxvit: Multi-axis vision transformer.
\newblock In \emph{European conference on computer vision}, pages 459--479, 2022.

\bibitem[Ulyanov et~al.(2016)Ulyanov, Lebedev, Vedaldi, and Lempitsky]{b52}
Dmitry Ulyanov, Vadim Lebedev, Andrea Vedaldi, and Victor Lempitsky.
\newblock Texture networks: Feed-forward synthesis of textures and stylized images.
\newblock \emph{arXiv preprint arXiv:1603.03417}, 2016.

\bibitem[Vaswani(2017)]{b18}
A Vaswani.
\newblock Attention is all you need.
\newblock \emph{NeurIPS}, 2017.

\bibitem[Wang and Liu(2024)]{b25}
Zijia Wang and Zhi-Song Liu.
\newblock Stylemamba: State space model for efficient text-driven image style transfer.
\newblock \emph{arXiv preprint arXiv:2405.05027}, 2024.

\bibitem[Wang et~al.(2023)Wang, Zhao, and Xing]{b14}
Zhizhong Wang, Lei Zhao, and Wei Xing.
\newblock Stylediffusion: Controllable disentangled style transfer via diffusion models.
\newblock In \emph{ICCV}, pages 7677--7689, 2023.

\bibitem[Wei et~al.(2022)Wei, Deng, Tang, Pan, and Dong]{b55}
Hua-Peng Wei, Ying-Ying Deng, Fan Tang, Xing-Jia Pan, and Wei-Ming Dong.
\newblock A comparative study of cnn-and transformer-based visual style transfer.
\newblock \emph{Journal of Computer Science and Technology}, 37\penalty0 (3):\penalty0 601--614, 2022.

\bibitem[Wright and Ommer(2022)]{b45}
Matthias Wright and Bj{\"o}rn Ommer.
\newblock Artfid: Quantitative evaluation of neural style transfer.
\newblock In \emph{DAGM GCPR}, pages 560--576, 2022.

\bibitem[Yang et~al.(2025)Yang, Zhou, Sun, Li, Liu, Lu, Ma, and Yan]{b68}
Hao Yang, Qianyu Zhou, Haijia Sun, Xiangtai Li, Fengqi Liu, Xuequan Lu, Lizhuang Ma, and Shuicheng Yan.
\newblock Pointdgmamba: Domain generalization of point cloud classification via generalized state space model.
\newblock In \emph{Proceedings of the AAAI Conference on Artificial Intelligence}, 2025.

\bibitem[Yao et~al.(2019{\natexlab{a}})Yao, Ren, Xie, Liu, Liu, and Wang]{b56}
Yuan Yao, Jianqiang Ren, Xuansong Xie, Weidong Liu, Yong-Jin Liu, and Jun Wang.
\newblock Attention-aware multi-stroke style transfer.
\newblock In \emph{Proceedings of the IEEE/CVF conference on computer vision and pattern recognition}, pages 1467--1475, 2019{\natexlab{a}}.

\bibitem[Yao et~al.(2019{\natexlab{b}})Yao, Ren, Xie, Liu, Liu, and Wang]{b6}
Yuan Yao, Jianqiang Ren, Xuansong Xie, Weidong Liu, Yong-Jin Liu, and Jun Wang.
\newblock Attention-aware multi-stroke style transfer.
\newblock In \emph{CVPR}, pages 1467--1475, 2019{\natexlab{b}}.

\bibitem[Yin et~al.(2024)Yin, Li, and Dong]{b34}
Zhuowen Yin, Chengru Li, and Xingbo Dong.
\newblock Video rwkv: Video action recognition based rwkv.
\newblock \emph{arXiv preprint arXiv:2411.05636}, 2024.

\bibitem[Yu(2015)]{b39}
F Yu.
\newblock Multi-scale context aggregation by dilated convolutions.
\newblock \emph{arXiv preprint arXiv:1511.07122}, 2015.

\bibitem[Yuan et~al.(2024)Yuan, Li, Qi, Zhang, Yang, Yan, and Loy]{b32}
Haobo Yuan, Xiangtai Li, Lu Qi, Tao Zhang, Ming-Hsuan Yang, Shuicheng Yan, and Chen~Change Loy.
\newblock Mamba or rwkv: Exploring high-quality and high-efficiency segment anything model.
\newblock \emph{arXiv preprint arXiv:2406.19369}, 2024.

\bibitem[Zhang et~al.(2018)Zhang, Isola, Efros, Shechtman, and Wang]{b43}
Richard Zhang, Phillip Isola, Alexei~A Efros, Eli Shechtman, and Oliver Wang.
\newblock The unreasonable effectiveness of deep features as a perceptual metric.
\newblock In \emph{CVPR}, pages 586--595, 2018.

\bibitem[Zhang et~al.(2022{\natexlab{a}})Zhang, Li, Li, Jia, and Zhang]{b10}
Yabin Zhang, Minghan Li, Ruihuang Li, Kui Jia, and Lei Zhang.
\newblock Exact feature distribution matching for arbitrary style transfer and domain generalization.
\newblock In \emph{CVPR}, pages 8035--8045, 2022{\natexlab{a}}.

\bibitem[Zhang et~al.(2022{\natexlab{b}})Zhang, Tang, Dong, Huang, Ma, Lee, and Xu]{b46}
Yuxin Zhang, Fan Tang, Weiming Dong, Haibin Huang, Chongyang Ma, Tong-Yee Lee, and Changsheng Xu.
\newblock Domain enhanced arbitrary image style transfer via contrastive learning.
\newblock In \emph{ACM SIGGRAPH}, pages 1--8, 2022{\natexlab{b}}.

\bibitem[Zhang et~al.(2023)Zhang, Huang, Tang, Huang, Ma, Dong, and Xu]{b13}
Yuxin Zhang, Nisha Huang, Fan Tang, Haibin Huang, Chongyang Ma, Weiming Dong, and Changsheng Xu.
\newblock Inversion-based style transfer with diffusion models.
\newblock In \emph{CVPR}, pages 10146--10156, 2023.

\bibitem[Zhou et~al.(2023)Zhou, Li, He, Yang, Cheng, Tong, Ma, and Tao]{b70}
Qianyu Zhou, Xiangtai Li, Lu He, Yibo Yang, Guangliang Cheng, Yunhai Tong, Lizhuang Ma, and Dacheng Tao.
\newblock Transvod: end-to-end video object detection with spatial-temporal transformers.
\newblock \emph{IEEE Transactions on Pattern Analysis and Machine Intelligence}, 45\penalty0 (6):\penalty0 7853--7869, 2023.

\bibitem[Zhu et~al.(2024)Zhu, Liao, Zhang, Wang, Liu, and Wang]{b36}
Lianghui Zhu, Bencheng Liao, Qian Zhang, Xinlong Wang, Wenyu Liu, and Xinggang Wang.
\newblock Vision mamba: Efficient visual representation learning with bidirectional state space model.
\newblock \emph{arXiv preprint arXiv:2401.09417}, 2024.

\end{thebibliography}
}

\end{document}